\documentclass[journal,web]{IEEEtran}
\usepackage{url,lineno,microtype,subcaption}
\usepackage{times}
\usepackage{epsfig}
\usepackage{graphicx}
\usepackage{amsmath}
\usepackage{amssymb}
\usepackage{algorithm}
\usepackage{algorithmicx}
\usepackage{algpseudocode}
\usepackage{bm}
\usepackage{color}
\usepackage{multirow}
\usepackage{makecell}
\usepackage{url}
\usepackage{array}
\usepackage{amsmath,amssymb,amsfonts}
\usepackage{bm}
\usepackage{bbding}
\usepackage{booktabs}
\usepackage{cite}
\usepackage{epsfig}
\usepackage{graphicx}
\usepackage{makecell}
\usepackage{multirow}
\usepackage{setspace}
\usepackage{times}
\usepackage{textcomp}
\usepackage{url}
\usepackage{verbatim}
\usepackage{xr}
\usepackage[pagebackref,breaklinks,colorlinks]{hyperref}
\usepackage{subfloat} 
\usepackage{booktabs} 
\usepackage{threeparttable} 
\usepackage[capitalize]{cleveref}
\crefname{section}{Sec.}{Secs.}
\Crefname{section}{Section}{Sections}
\Crefname{table}{Table}{Tables}
\crefname{table}{Tab.}{Tabs.}
\newtheorem{Hypothesis}{\bf Hypothesis}

\title{Rethinking Pretraining as a Bridge  \\ from ANNs to SNNs}

\author{Yihan Lin,$^{1}$, Yifan Hu,$^{1}$ Shijie~Ma,$^{2}$ Guoqi Li$^{2,*}$, Dongjie Yu,$^{3}$
\thanks{
 $^{1}$ Center for Brain-Inspired Computing Research, Tsinghua  University, Beijing, China.
 $^{2}$ Institute of Automation of Chinese Academy of Sciences, Beijing, China.
 $^{3}$ School of Vehicle and Mobility, Tsinghua  University, Beijing, China.
 The Corresponding authors: Guoqi~Li. (E-mail: liguoqi@mail.tsinghua.edu.cn).}%
}

\begin{document}
\maketitle
\markboth{In Review}  
{Lin \MakeLowercase{\textit{et al.}}: Rethinking Pretraining as \\ a Bridge from ANNs to SNNs}
\begin{abstract}
\noindent Spiking neural networks (SNNs) are known as a typical kind of brain-inspired models with their unique features of rich neuronal dynamics, diverse coding schemes and low power consumption properties. How to obtain a high-accuracy model has always been the main challenge in the field of SNN. Currently, there are two mainstream methods, i.e., obtaining a converted SNN through converting a well-trained Artificial Neural Network (ANN) to its SNN counterpart or training an SNN directly. However, the inference time of a converted SNN is too long, while SNN training is generally very costly and inefficient. In this work, a new SNN training paradigm is proposed by combining the concepts of the two different training methods with the help of the pretrain technique and BP-based deep SNN training mechanism. We believe that the proposed paradigm is a more efficient pipeline for training SNNs. The pipeline includes pipe-S for static data transfer tasks and pipe-D for dynamic data transfer tasks. SOTA results are obtained in a large-scale event-driven dataset ES-ImageNet. For training acceleration, we achieve the same (or higher) best accuracy as similar LIF-SNNs using 1/10 training time on ImageNet-1K and 2/5 training time on ES-ImageNet and also provide a time-accuracy benchmark for a new dataset ES-UCF101. These experimental results reveal the similarity of the functions of parameters between ANNs and SNNs and also demonstrate the various potential applications of this SNN training pipeline.
\end{abstract}


\begin{figure}[!t]
    \centering
    \includegraphics[width=1 \linewidth]{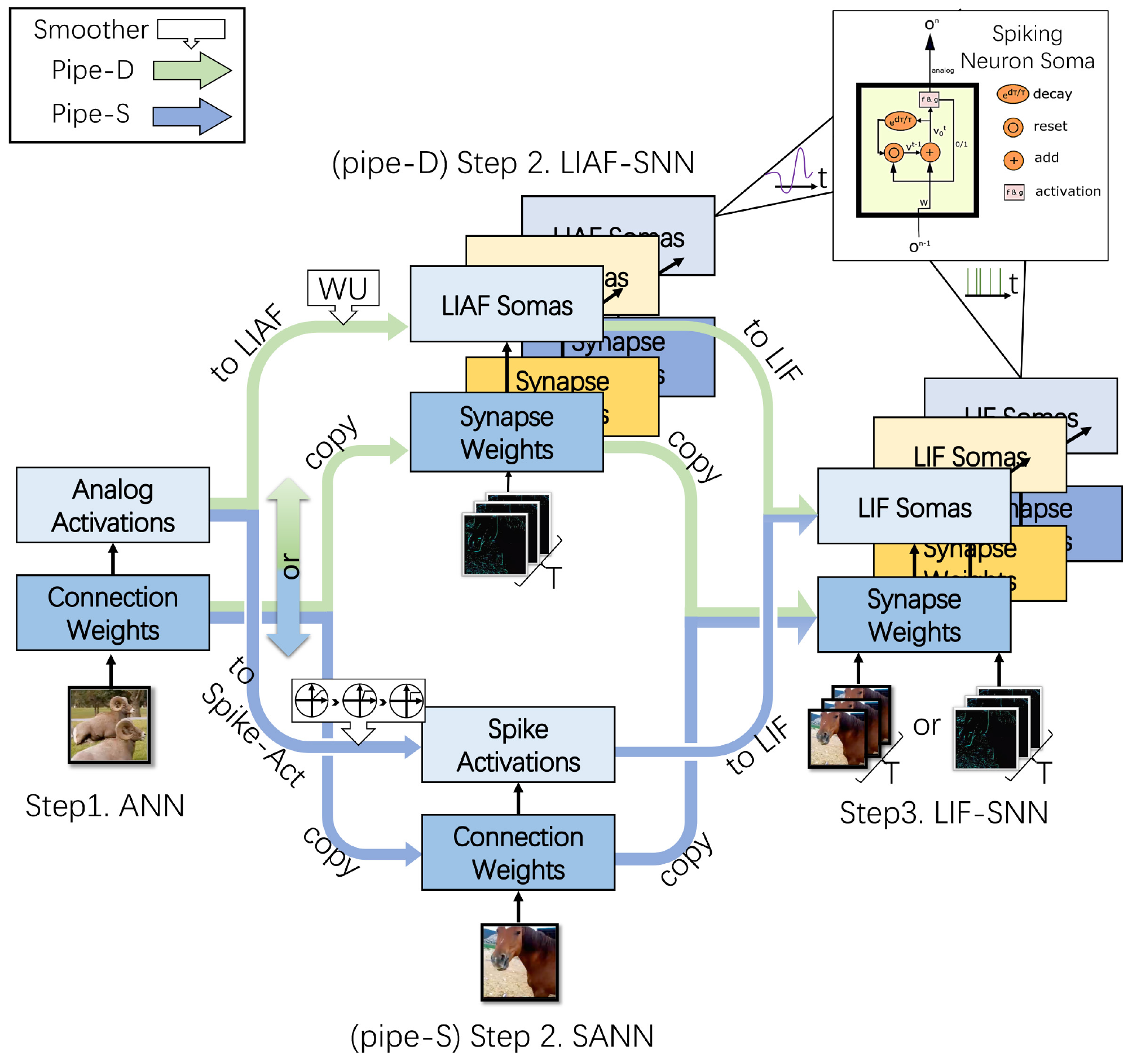}
    \caption{{\bf ANN-SNN transfer pipeline}. The training pipeline has two branches according to different datasets. When the target data is static data (like images), we prefer to use {\bf pipe-S}, while we need to train LIF-SNN on dynamic data (like videos), we use {\bf pipe-D}. Both of the sub-pipeline can be divided into three steps: training the original ANN model, fine-tuning on the transitional models (a spike-activation ANN or a LIAF-SNN \cite{wu2021liaf}) and fine-tuning on the LIF model.  Warmup (WU) step and sharpened ReLU are adopted as the smoothing methods, and the detailed process is introduced in the Method section.} 
    \label{fig:train}
\end{figure}

\section{Introduction}\label{sec:Introduction}
\label{sec:introduction}
The development of Neural Networks (NN) is a hot spot for both academia and industry. Although great success has proved the value of ANNs, they are divorced from the biological basis of real neural networks and depend heavily on high-performance computation. The appearance of brain-inspired SNNs fills in this blank. In 1997, spiking neuron networks based on Leaky Integrate-and-Fire(LIF) were proposed \cite{SpikingNeuron} as the third generation of NN. The modeling of spiking neurons imitates the propagation mechanism of the electrical signal of biological neurons using spikes to transmit information. Spiking neurons have a solid biological basis and spatiotemporal information coding complexity. SNNs are good at the event-driven and energy-constrained tasks\cite{TSMINC,TIANJI}. An increasing amount of new training algorithms based on different degrees of biological plausibility have also been developed. Due to the progress, SNNs have come to the public view and become a new road towards artificial general intelligence. 

Although SNNs can simulate real neurons and show extraordinary low power consumption, the directly-trained SNN models \cite{STBP} are often too small in scale limited by training algorithms and long training time. The small-scale networks are often of poor performance and hard to be applied in the real-world  applications with huge amount of data. Some algorithms provide the solution works on specific network structures \cite{Zheng_Wu_Deng_Hu_Li_2021,fang2021deep}, which also have limitation on applications. So a more general algorithm for obtaining large-scale SNNs efficiently is urgently needed.

Back to the motivation of SNNs, researchers hope that SNNs could provide a nice simulation of biological neural networks. Although the powerful training algorithms endow SNNs with learning ability, the critical problem is that the biological instinct, which is brought by the prior knowledge stored in deoxyribonucleic acid (DNA) to recognize the world, is not reflected by SNNs. The pretrain techniques inspire us, as many existing well-trained ANN models provide a large amount of prior knowledge. It is also well known that pretraining is an important technique to increase the scale of the model and accelerate the training process. At present, most of the commonly used SNNs have incorporated the parameter structures of ANNs, so researchers have made efforts to reuse the weights of a well-trained ANN to obtain a converted-SNN without training\cite{o2013real}. The performance of converted-SNNs is close to the original ANNs, but they are not widely used due to the data constraints (only suitable for static data) and long inference time. In computer vision tasks, it has been proved that an ANN can help to train an SNN with a similar structure by reusing the weights and low-level features\cite{wu2019tandem}. Inspired by the results, we design a training pipeline with two branches named {\bf pipe-D} (pipe-dynamic) and {\bf pipe-S} (pipe-static) depending on the type of data. We prefer to use {\bf pipe-S} on static-static data transferring tasks and {\bf pipe-D} on dynamic-data relative transferring tasks. These pipelines are further illustrated in {\bf Fig. \ref{fig:train}}. Leaky-Integrate-and-Analog-Fire (LIAF) SNN \cite{wu2021liaf} and Spiking-activation-ANN (SANN) are introduced as the intermediaries of parameter transition. LIAF SNN is a variant model of LIF-SNN, whose amplitude of the spikes is represented by continuous value, so it does not suffer from the inaccurate gradient during training as LIF-SNN. The SANN also has the common characteristics of both LIF-SNNs and ANNs and makes the transition of parameters smoother. 

We believe that the pretrained parameters will help SNN for difficult tasks, even when the distributions and modes of the data change. Moreover, our experiments prove that such a training pipeline can significantly facilitate the training of SNNs, accelerate convergence, and improve generalization performance without any other tricks. 

The main contributions of this work are threefold. 

{\bf (i)} A new training pipeline of SNNs is proposed in this paper, which uses pretrained ANN models to accelerate the training of SNNs and improve SNNs' performance. To the best of our knowledge, it is the first attempt to propose a complete pretrain-finetune framework for training LIF-SNNs and it would become a bridge from ANNs to SNNs, further promoting the application of SNNs. 

{\bf (ii)} Comprehensive experiments have been conducted, providing evidence to support that (1) Using the same training methods, SNNs and ANNs will grow to have similar feature extraction and inference abilities. (2) Similar feature extraction and inference abilities may help both ANNs and SNNs to gain good inference abilities.

{\bf (iii)} We obtain many high-performance SNN models while saving several times of the training cost in various tasks. SOTA results are provided in this paper on ES-Imagenet (52.25\%/43.74\% for LIAF/LIF-SNN) and we also obtain competitive test accuracy on ImageNet-1K, CIFAR10-DVS and ES-UCF101 with significant training acceleration. ES-UCF101 is a new lightweight event stream video classification dataset we propose for long-time sparse spatial-temporal feature extraction.

\section{Related Work}

\subsection{Pretrain Technique and Transfer Learning}

Pretrain techniques are often applied in situations where large amounts of generalized data are available, but task-specific domain data is few and difficult to support practical training. This technique significantly reduces the training time on specific tasks and improves performance. When the pretrain and finetune process are conducted on a different dataset, we call it transfer learning (TL). In this paper, we adopt the ideas from feature-based TL. In CV tasks, the external image data can be used to obtain a feature extractor for neural networks initialization or layer transfer to the target task. Because the low-level features of images are reusable, the optimization of the target task can be facilitated by reusing the learned low-level features from other datasets like ImageNet \cite{Imagenet}. Jason et al. investigated the effects of freezing parameters of different layers on the transferability and representation ability \cite{imagenet_transfer_14}. In tasks with similar features, adversarial optimization is often used to extract shared features between source and target tasks by removing domain properties, which enhances the robustness \cite{ganin2016adversarial}. 

\subsection{SNN Training Algorithms}

ANN-to-SNN conversion can be regarded as the primary method of transferring ANN pretrained models to SNNs. The early conversion methods use the mean-firing-rate approximation of LIF neurons to approximate probabilities during training\cite{o2013real} and many normalization methods are proposed for better conversion \cite{diehl2015fast,spikingYolo}. Nowadays the advanced conversion methods can equip a converted-SNN with performance comparable to the corresponding ANN. But these methods can only be used in static data. The long inference time also prevents the converted SNNs from wide applications. The direct training method is another way to get an SNN, where spatio-temporal backpropagation (STBP) \cite{STBP} is a frequently-used BP-based method. STBP uses the gradient approximation to avoid the discontinuity of the spiking activation function and apply backpropagation through time (BPTT) \cite{BPTT1990} on SNNs. But the inaccurate gradient propagation in the BP-based method makes it difficult to obtain a high-performance LIF-SNN, and the multi-time-step data flow slows down the process of gradient propagation, leading to an unbearable long training time.

In order to obtain high-performance SNNs more efficiently in various scenarios, we consider adopting STBP and the pretrain technique jointly. There are some early explorations in this field. Wu et al. \cite{wu2019tandem} provides a method that guides the BP process by the gradient of ANN and some spiking characteristics of SNN. Hybrid neural state machine (H-NSM) applies the ANN-SNN hybrid architectures, where the SNN and ANN are trained together using various rules under different conditions\cite{NHSM-lp}. However, these attempts are simple or of limited applications. Rathi et al. provide a more inspiring method that uses Spike Timing Dependent BP to finetune the model based on a converted SNN\cite{Rathi2020Enabling}, but it is still limited in applicable scenarios and performance, which urges us to propose a more universal and efficient ANN-to-SNN pretraining framework.

\section{Method}

\subsection{Models}

The LIF-SNN model we used is simplified from the continuous LIF model \cite{SpikingNeuron}. LIF model considers the influence of leaky current of neurons, where the membrane voltage can be described as 

\begin{equation}
    \label{LIF-connect}
    U(t+1)=U(t)e^{-\frac{1}{\tau}}+X(t).
\end{equation} 
 
Here, $U(t)$ denotes the membrane voltage, $\tau$ denotes a time coefficient, $X(t)$ is the weighted input current. When a neuron fires a spike, its membrane will be reset to 0. To construct a multi-layer SNN in computer, we define the discrete version of the LIF model in the $l^{th}$ layer as

\begin{eqnarray}
     \label{Eq:LIF-1}
     & X^{l}(t)=W^l * O^{l-1}(t),\\
     & U^{l}(t+1)=U^{l}(t) e^{-\frac{1}{\tau}}(1-O^{l}(t))+X^{l}(t) + \beta, \\
     & O^{l}(t+1)=F(U^{l}(t+1) - U_{th}),
\end{eqnarray}
where $W^l$ is the weight matrix, $U_{th}$ denotes threshold voltage and $O(t)$ is spike train at timestamp $t$. $X$ is the input spike train, for example, $\underbrace{\{1,0,0,...,1,0,1\}}_{T}$ and $T$ is the simulation time. Rate Coding is often used in LIF-SNN, where we will calculate the average fire rate of neurons in the output layers. $F$ is the spike activation function and we often use step function as $F$ like:

\begin{equation}
    \label{LIF-spikefun}
    O^{l}(t)=F(U^{l}(t))=
    \begin{cases}
    1, & U^{l}(t) > U_{th} \\
    0, &  others.
    \end{cases}
\end{equation}

In LIAF models \cite{wu2021liaf}, the spike activation function $F$ will have continuous output, while the membrane reset signal is still binary. 

\subsection{Training Pipelines}

For different datasets, we design different pipelines for the pretrain-finetune process.

(1) {\bf pipe-S:} (ANN $\rightarrow$ SANN $\rightarrow$ LIF)  When the target dataset is a static image dataset, we train an ANN on this dataset, then fine-tune the ANN with spiking-activation. After that, the weights from the source SANN are copied to the target SNN and get fine-tuned with the extension of the temporal dimension.

(2) {\bf pipe-D:} (ANN $\rightarrow$ LIAF-SNN$\rightarrow$ LIF-SNN) For the dynamic datasets like DVS-dataset or video dataset, we can adopt DVS-to-gray reconstruction method (without extra data) or an additional static source dataset to obtain a pretrained ANN model, then fine-tune a LIAF-SNN model on the target dynamic dataset. Finally, we fine-tune the LIF-SNN from the LIAF version. 
 
In brief, the training pipeline is based on a pretrained ANN model and introduces the spiking-activation and the iterative neuron mechanism serially. {\bf pipe-D} is more general and also suitable for static datasets, and {\bf pipe-S} is faster, but it is unfriendly to dynamic tasks. The main idea of the training pipeline is established based on these hypotheses:

\begin{Hypothesis}
Using the same train methods, SNNs and ANNs share similar feature extraction-inference abilities. 
\label{Hypothesis:1}
\end{Hypothesis}

\begin{Hypothesis}
On the same tasks, similar feature extraction and inference ability may boost both ANNs and SNNs. 
\label{Hypothesis:2}
\end{Hypothesis}

Experiments are designed to verify {\bf Hypothesis \ref{Hypothesis:1} and  \ref{Hypothesis:2}}. We will firstly compare the difference among the directly trained ANN, LIAF-SNN, and LIF-SNN with the same BP-based training process, then visualize their weights. If {\bf Hypothesis \ref{Hypothesis:1}} holds, similar patterns of weights should be observed, which lead to similar features for the same samples. For {\bf Hypothesis \ref{Hypothesis:2}}, the training pipeline will be examined on different tasks using both pipelines, expecting the validation results and convergence speeds of SNNs to be significantly improved. STBP \cite{STBP} algorithm is implemented for training deep SNNs, which is similar to the BPTT algorithm for ANNs. The basic ResNet18\cite{ResNet} and LRCN \cite{donahue2015long} are chosen as backbones of models. 

\subsection{Transfer Process Smoothing}

For both {\bf pipe-D} and {\bf pipe-S}, we have introduced different smoothing methods when the activation is changed or when some of the neural layers are changed. These smooth methods increase the efficiency of the pipeline. 

In view of the potential obstacle caused by the non-differential spiking activation function in {\bf pipe-S}, we incorporate the second step into the first one and offer a smoother procedure to obtain an SANN. A progressively sharpened ReLU (SReLU) \cite{severa2019training} is adopted as
 \begin{equation}
     F(U) = \begin{cases}
0, & U < \alpha \\
1, & U > \beta\\
\frac{U-\alpha}{\beta-\alpha}, & others
\end{cases},
 \end{equation}
and we assert that $\alpha < \beta$ and $\alpha + \beta=2U_{th}$. As $\alpha$ increases linearly from $0$ to $U_{th}$ with training epochs, the function will approach the spiking activation in the end, while its derivative is set unchanged as
\begin{equation}
    F'(U)=\begin{cases}
    1, & 0\le U \le 2U_{th}\\
    0, & others
    \end{cases}
\end{equation}
to offer a stable convergence. The spiking model learns how to express effectively when the temporal dimension is extended at the final step, which is of vital importance to avoid the lengthy simulation in ANN-converted SNNs.

For transfer tasks that network structures are changed (where we often use {\bf pipe-D}), it is recommended to add several warmup (WU) epochs (and usually one epoch is enough) between step 1 and step 2. During the WU epochs, we freeze the parameter of the unchanged layers and only retrain the changed layers from scratch, whose parameters cannot be transferred directly. The changed layers are usually the first layer and the last layer, which are directly related to the size of input data and output predictions. Without WU epochs, the noise information brought by the randomly initialized layers may disturb the well-initialized layers, which sometimes disrupts the pretrain-finetune process.

\begin{figure*}[!t]
    \centering
    \includegraphics[width=1 \linewidth]{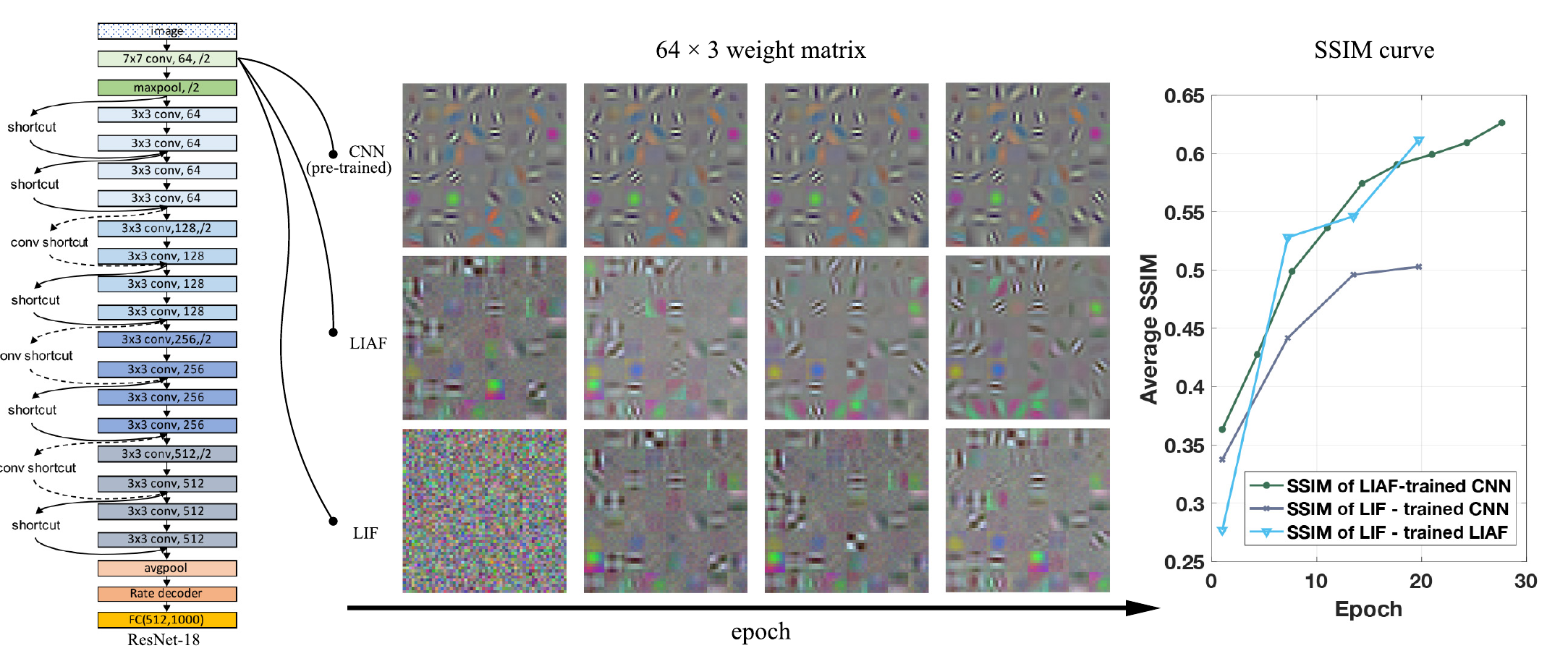}
    \caption{{\bf Weight maps of CNN, LIAF-SNN, and LIF-SNN in ImageNet experiment in {\bf pipe-D}}.  We place these convolution kernels in the order of $OutChannels$, where each $7 \times 7$ square contains $[i^{th}OutChannels,:,:,:]$. The color of the weight presents the absolute value of the weight, which may be influenced by the membrane update mechanism of the LIF/LIAF neuron in the training process. The orders have been adjusted by using the Hungarian algorithm to achieve the maximum matching, where SSIM is used to measure the similarity of weights. The SSIM curve indicates that with the performance of LIAF-SNN increasing, its parameter is glowingly similar to CNN's, While LIF-SNN gradually evolves to be more like LIAF-SNN. Here we use an SGD optimizer without momentum in our experiments to keep the consistency of the training process.}
    \label{fig:Imagenet_param}
\end{figure*}

\section{Experiments}

We use three types of datasets in our experiments. They are traditional RGB dataset ImageNet-1K\cite{Imagenet}, static-data-converted/recorded dynamic vision sensor (DVS) dataset ES-ImageNet \cite{lin2021esimagenet} and CIFAR10-DVS\cite{dvscifar10}, and dynamic-data-converted ES-UCF101. They are different in the form of data, especially the encoding mechanisms. The latter three datasets share similar data coding characteristics with SNNs' spiking data flow. We will have extensive experiments on those datasets and compare the classification accuracy with the SOTA results of SNNs.    

\subsection{Exp. 1: Static-Static Experiments}

ImageNet-1k (or \emph{ILSVRC2012})\cite{ILSVRC15} is a well-known large-scale visual dataset. It contains about 1.3M image samples collected from the real scene with 1000 different categories. ResNet-18 is selected as our backbone, and we test both pipelines and we set $T=6$ for the SNNs.

\begin{table}[htb]
\centering
\caption{Experiment Results (Top-1 Accuracy) on ImageNet.}
\begin{threeparttable}
\resizebox{\linewidth}{!}{
\begin{tabular}{m{3cm}m{2cm}<{\centering}m{1cm}<{\centering} m{1.5cm}<{\centering}}
\toprule
 Network Structure\tnote{a}  &  Pretrain Method\tnote{b} & Acc. ($1^{st}$ Epoch) & Best Acc. (Epoch) \\ 
 \hline
LIF-R18  &  |     &  12.31& 62.10(99)     \\
SEW-R18\cite{fang2021deep}                  &  |     & |     & 63.18(300+)   \\
tdBN+LIF-R34\cite{Zheng_Wu_Deng_Hu_Li_2021} &  |     & |     & 63.72 \\
Convert-LIFR34\cite{Rathi2020Enabling}        &  |     & |     & 61.48(T=250)\\
Convert-IF-R34\cite{10.3389/fnins.2019.00095}        &  |     & |     & 65.47(T=2500) \\
spiking-VGG15 \cite{10.3389/fnins.2020.00535}        &  |     & |     & 66.56 \\
DIETE-VGG16 \cite{rathi2021dietsnn}        &  |     & |     & 69.00(T=5) \\
\hline
SCNN-R18  &  pipe-S step1  &    |     & 50.25       \\
LIAF-R18  &  pipe-D step1  &  49.83   & 64.04     \\
\hline
LIF-R18 &  pipe-S step2 &  58.08   & {\bf 62.81(30)}\\
LIF-R18 &  pipe-D step2 &  48.27   & 60.32(32)      \\
\bottomrule
\end{tabular}}
\begin{tablenotes}
    \footnotesize
    \item[a]  R18/R34 denoted basic ResNet-18/34 structure.
    \item[b]  SReLU and WU are adopted for pipe-S and pipe-D separately.
    \item[] The pretrained ResNet-18 is with 64.1\% Acc.
\end{tablenotes}
\end{threeparttable}
\label{table:result-ImageNet}
\end{table}

{\bf Training Result.} At the beginning of the pretraining stage, the direct transplantation of the convolutional neural network's (CNN's) weights to its SNN counterpart is of little help, but as the activation evolves into the step function while being finetuned, the corresponding CNN grows into a well-behaved Spiking-activation CNN (SCNN). When making inference with strict spiking activation, the model and its parameters obtain a top-1 accuracy of 50.25\% on ImageNet. Then surprisingly, the accuracy increases to 58.08\% after only one epoch of finetuning with extended temporal dimension, which indicates the importance of further adapting the pretrained model to spatial-temporal dynamics. After the subsequent decay of the learning rate, it turns out that only one-tenth of the training epochs are required for the finetune stage at its fastest pace, to achieve the same accuracy as that of the model trained from scratch. A training procedure (e.g. of 30 epochs) with moderate learning rate decay will yield a higher accuracy, as in {\bf Table \ref{table:result-ImageNet}}. Through pretraining to learn the extraction ability of spatial features and finetuning to learn effective representation in spatial-temporal dynamics, our {\bf pipe-S} provides a much faster and effective way to train an SNN. More details can be found in supplementary materials. We also test {\bf pipe-D} and it also shows acceleration effect and boosts the training of LIF-SNN. However, for static tasks, the redundancy of {\bf pipe-D} is somehow harmful for gradient propagation and leads to a slightly worse result.

{\bf Weight Analysis.} Usually, the shallow layers of the CNNs often focus on low-level features. In contrast, deeper layers in the CNNs are closely related to the selected dataset and task \cite{imagenet_transfer_14}. So we focus on the output features of the shallow layers to judge whether they have the potential to help with training through direct transplantation of the parameters. Because the number of input channels of the convolution kernel in the first layer is $3$, We can treat them as $N$ 3-channel $7 \times 7$ RGB images according to the order of output channels. Structural Similarity Index Measure (SSIM) is used to measure the similarity between different weights, which is defined as

 \begin{eqnarray}
 &L(x,y)=\frac{2\mu_x\mu_y+C_1}{\mu_x^2+\mu_y^2+C_1},\\
 &C(x,y)=\frac{2\sigma_x\sigma_y+C_2}{\sigma_x^2+\sigma_y^2+C_2},\\
 &S(x,y)=\frac{\sigma_{xy}+C_3}{\sigma_x\sigma_y+C_3},\\
 &SSIM(x,y)= L(x,y)C(x,y)S(x,y),
 \end{eqnarray}
 where $\mu_x$ and $\sigma_x$ are the mean value and the unbiased standard deviation of the data $x$ respectively. $L(x,y), C(x,y), S(x,y)$ measure the similarity between two images by luminance, contrast, and structure. With the help of SSIM, we can establish a table of similarity degrees of weights between two different networks. We train ResNet-18, LIAF-ResNet-18 and LIF-ResNet-18 using the same set of initial parameters, algorithm, random seed, and hyper-parameters under the same training environment. Hungarian algorithm is used here to find the optimal matching result to verify whether the weights of the network are similar from a proper perspective. After the matching process,  weights are visualized in the paired order, as shown in {\bf Fig. \ref{fig:Imagenet_param}}. During training, the weight patterns of the SNN's convolution kernels are evolving to the final patterns of those of the CNN. For a more rigorous analysis, we also draw the SSIM curve of SNNs during the training progress, which also supports our hypothesis. 

\subsection{Exp. 2: Static-Dynamic Experiments}

\begin{table}[htb]
\centering
\caption{Experiment Results (Top-1 Accuracy) on ES-ImageNet. }
\begin{threeparttable}
\resizebox{\linewidth}{!}{%
\begin{tabular}{m{2.2cm}m{3.7cm}<{\centering}m{0.7cm}<{\centering}m{1cm}<{\centering}}
\toprule
  Network\tnote{a}   & Pretrain Method\tnote{b}   & Acc. ($1^{st}$ Epoch) & Best Acc. (Epoch) \\ 
 \hline
LIAF-R18\cite{lin2021esimagenet} &  |     &  |&  42.54(50)\\
LIAF-R34\cite{lin2021esimagenet} &  |     &  |&  47.47(50)\\
LIF-R18\cite{lin2021esimagenet}  &  |  & | &  39.89(50)  \\
LIF-R34\cite{lin2021esimagenet}  &  | &  | &  43.42(50)   \\
\hline
LIAF-R18 & pipe-D step1 without WU   &  14.33   &  46.06(11)  \\
LIAF-R18$^{G}$ & pipe-D step1 &  19.74   &  50.54(16)  \\
LIAF-R18$^{R}$ & pipe-D step1   &  29.92   &  {\bf 52.25(16)}\\
\hline
LIF-R18  & pipe-D step2 + LIAF-18$^{G}$  & 15.60  & {\bf 43.47(11)}\\
LIF-R18  & pipe-D step2 + LIAF-18$^{R}$  & 17.68  & {\bf 43.74(18)}\\
\bottomrule
\end{tabular}}
\begin{tablenotes}
    \footnotesize
    \item[a]LIAF-R18$^{G}$: pretrained ANN obtained on the reconstructed Gray 
    \\ dataset from ES-ImageNet with $Acc = 42.94\%$.
    \\ LIAF-R18$^{R}$: pretrained ANN obtained {\bf Exp. 1}.
    \item[b]  WU are adopted for pipe-D.
\end{tablenotes}
\end{threeparttable}
\label{table:result-ES18}
\end{table}

ES-ImageNet is an event-stream (ES) classification dataset converted from ImageNet-1K, but their data modes are different. The samples in ES-ImageNet are sequences data of quads $(x,y,t,p)$, where $(x, y)$ is the topological coordinates of the pixel, $t$ is the time of spike generation, and $p$ is the ternary polarity of the spike $\in\{-1,0,1\}$. We usually use a two channel event frame format to structure these data to form event frame stream data, which is similar to video data. The ES-ImageNet consists of about 1.3M ES samples, with each sample having 8 time steps and about a 5\% none-zero rate. Its former SOTA result is 47.47\% obtained by a LIAF-SNN with ResNet-34 backbone\cite{lin2021esimagenet}. CIFAR10-DVS is another dynamic dataset, which is converted from CIFAR10 by using a DAVIS camera \cite{DAVIS} to record image motions on an LCD monitor\cite{dvscifar10}. The samples are also organized as $(x,y,t,p)$ quads. The data distribution of CIFAR10-DVS is much different from that of ImageNet or ES-ImageNet. The pipeline would be tested on CIFAR10-DVS using the pretrained models obtained on ImageNet-1K or ES-ImageNet, hoping that the benefits of the pipeline can also be observed on transfer tasks with a larger gap. We choose $T=8$ for pipe-D in these experiments, while $T$ is extended to $20$ in a case of the CIFAR10-DVS experiment.

\begin{figure}[!t]
    \centering
    \includegraphics[width=1 \linewidth]{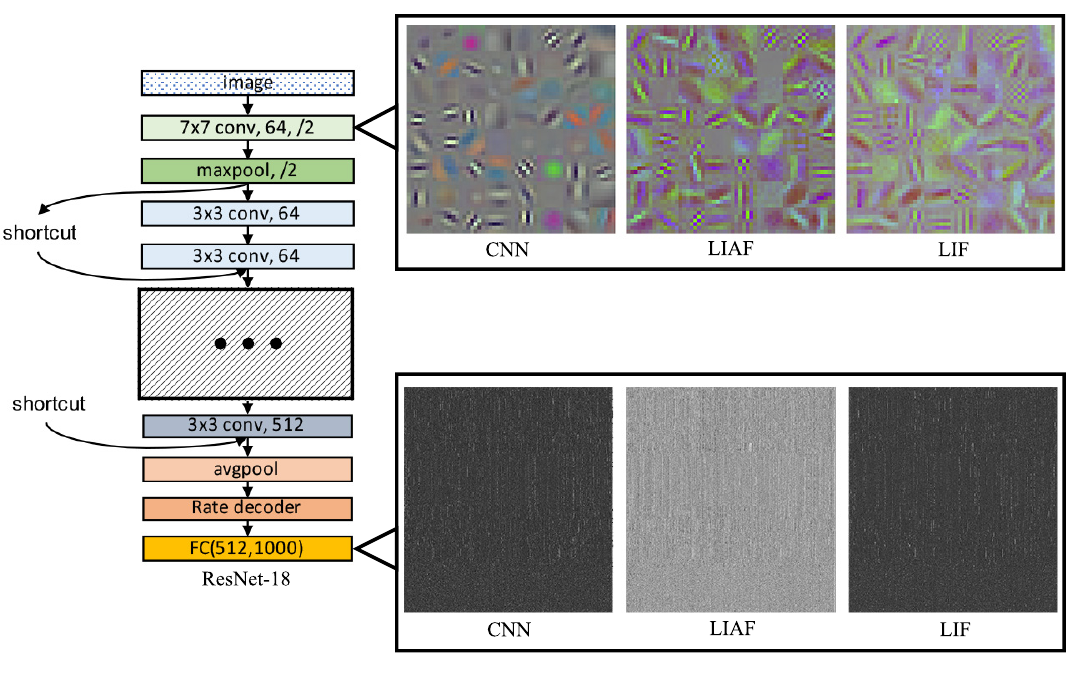}
    \caption{{\bf Weight on ES-ImageNet}. After the WU epochs, the first convolution layer on LIAF-ResNet-18 is retrained and the parameter matrices still show similar patterns as CNN with different input channels. At the same time, the highest-level embedding features also have similarities in the distribution of the weights of each prediction.} 
    \label{fig:ES_Imagenet_weight}
\end{figure}


%

\begin{table}[htb]
\centering
\caption{Transfer learning results on CIFAR10-DVS.}
\begin{threeparttable}
\resizebox{\linewidth}{!}{\begin{tabular}{l<{\centering}l<{\centering}m{1cm}<{\centering}}
\toprule
  Network & Pretrain Method   &  Acc.\\
\hline
\multicolumn{2}{l}{GCN \cite{bi2019graph}} & 54.00 \\
\multicolumn{2}{l}{LIF-SCNN \cite{STBP}} & 60.50 \\
\multicolumn{2}{l}{tdBN-RseNet18\cite{Zheng_Wu_Deng_Hu_Li_2021}} & 67.80 \\
\multicolumn{2}{l}{PLIF \cite{fang2021incorporating}(T=20)} & 74.80 \\
\hline
LIF-R18\tnote{-} & From scratch               & 53.64 \\
LIF-R18\tnote{-} & pipe-D (reuse ImageNet Exp.)   & 62.89 \\
LIF-R18\tnote{-} & pipe-D (reuse ES-ImageNet Exp.) & 66.02\\
\hline
LIF-R18\tnote{=} & From scratch          &    64.17 \\
LIF-R18\tnote{=} & pipe-D (reuse ES-ImageNet Exp.) & {\bf 70.52}\\
LIF-R18\tnote{+} & pipe-D (reuse ES-ImageNet Exp.) & {\bf 72.50}\\
\bottomrule
\end{tabular}}
\begin{tablenotes}
    \footnotesize
    \item[-] Event frames are obtained by down-sampling to $42\times42$.
    \item[=] Event frames keep the resolution as original $128\times128$.
    \item[+] $T$ is extended from 8 to 20.
\end{tablenotes}
\end{threeparttable}
\label{table:result-DVS-CIFAR10}
\end{table}

{\bf Training Result} We conduct an ablation experiment on the dataset with {\bf pipe-D}. It should be noted that, because there is a change in the number of the data's channel (RGB frame is three and event frame is two), we introduce a WU step before fine-tuning LIAF-SNN, where we only train the first layer of the network. The training results on ES-ImageNet with ResNet-18 structure can be found in {\bf Table \ref{table:result-ES18}}. The experimental results show that such a training pipeline is also effective in some cross-mode tasks, and with external data the network can obtain the highest accuracy. Because the cost of training SNN (especially the time) is several times to tens of times larger than that of ANN, the whole pipeline saves a lot of training cost compared with direct training. 

We then use the pretrain model obtained on ES-ImageNet and ImageNet-1K to apply {\bf pipe-D} on CIFAR10-DVS. The input size of the sample is resized to $42\times42$ by down-sampling for one case and keeping the original $128\times128$ for another. WU is used for the last layer, where the output dimension is changed from 1000 to 10. The results are shown in  {\bf Table \ref{table:result-DVS-CIFAR10}}. More details can be found in supplementary materials. 

{\bf Weight Analysis.} The parameter of three models (with the highest accuracy respectively) is visualized in {\bf Fig. \ref{fig:ES_Imagenet_weight}}, and we can still find similar patterns in the parameters of the first convolutional layer (although the number of input channels is different). That layer of LIAF is initialed randomly in step 2 of {\bf pipe-D}, and this shows that the weights of the deeper layers in the network may bring the information back to the new layers in the WU process. The feature maps of three different convolution layers are also visualized in {\bf supplementary materials Fig. S3}, where the feature maps of SNN are the average value of the output spike-frame sequences, i.e., the rate decoding process. 

\subsection{Exp. 3: Dynamic-Dynamic Experiments}

\begin{table}[htb]
    \centering
    \caption{{\bf LIAF results on ES-UCF101}, with \Checkmark for using pretrained models, while \XSolidBrush for not. For epoch columns, numbers in ( ) denote test accuracy($\%$) in corresponding epochs, for best Acc. column, numbers in ( ) denote corresponding epochs.}
    \label{table:video-LIAF-results}
    \begin{threeparttable}
    \resizebox{\linewidth}{!}{\begin{tabular}{ccccc}
    \toprule
    CNN-Enc & RNN-Dec & epoch-1\tnote{a} & epoch-2\tnote{b} & best Acc.(epoch)  \\
    \midrule
    \XSolidBrush & \XSolidBrush & 6(20.42) & 16(41.71) & 47.84(35) \\
    \Checkmark   & \XSolidBrush & 5(22.46) & 14(40.93) & \textbf{52.85}(56) \\
    \XSolidBrush & \Checkmark   & 4(21.35) & 16(41.41) & 47.84(51) \\
    \Checkmark   & \Checkmark   & \textbf{3(22.76)} & \textbf{11(41.32)} & 50.12(49) \\
    \bottomrule
    \end{tabular}}
    \begin{tablenotes}
        \footnotesize
        \item[a] \textbf{epoch-1} denotes the first epoch when test accuracy reaches $20\%$.
        \item[b] \textbf{epoch-2} denotes the first epoch when test accuracy reaches $40\%$.
    \end{tablenotes}
    \end{threeparttable}  
\end{table}
\begin{table}[!t]
    \centering
    \caption{{\bf LIF results on ES-UCF101}, with \Checkmark for using pretrained models at this step, while \XSolidBrush for not. 
    For epoch columns, numbers in ( ) denote test accuracy($\%$) in corresponding epochs, for best Acc. column, numbers in ( ) denote corresponding epochs. 
    }
    \label{table:video-LIF-results}
    \begin{threeparttable}
    \resizebox{\linewidth}{!}
    {\begin{tabular}
    {m{0.6cm}m{0.6cm}m{0.6cm}m{0.6cm}m{1.2cm}m{1.2cm}m{1cm}}
    \toprule
    \multicolumn{2}{c}{\bf Step 1} & \multicolumn{2}{c}{\bf Step 2} & \multicolumn{3}{c}{ }\\
    CNN Enc          & RNN Dec          & LIAF Enc         & RNN Dec          & epoch-1   & epoch-2   & best Acc.(epoch) \\
    \midrule
    \XSolidBrush & \XSolidBrush & \multicolumn{2}{c}{No step 2} & 11(20.00)  & ~~~~------   & 39.70(55)   \\
    \Checkmark   & \XSolidBrush & \multicolumn{2}{c}{No step 2} & 11(21.11) & 47(40.03) & 40.84(49)    \\
    \XSolidBrush & \Checkmark   & \multicolumn{2}{c}{No step 2} & ~~5(20.63)  & 26(40.36) & 45.14(54)    \\
    \Checkmark   & \Checkmark   & \multicolumn{2}{c}{No step 2} & ~~4(20.99)  & 18(41.56) & 46.94(56)    \\
    \hline
    \multicolumn{2}{c}{No step 1} & \Checkmark   & \XSolidBrush & ~~3(21.71)  & 10(40.60) & 47.72(60)    \\
    \multicolumn{2}{c}{No step 1} & \XSolidBrush & \Checkmark   & ~~3(21.65)  & 18(40.15) & 46.94(48)    \\
    \multicolumn{2}{c}{No step 1} & \Checkmark   & \Checkmark   & ~~1(35.32)  & ~~3(40.27)& 49.76(47)    \\
    \hline
    \Checkmark   & \XSolidBrush & \Checkmark   & \XSolidBrush & ~~3(23.81)  & 11(40.42) & 47.66(57)    \\
    \Checkmark   & \XSolidBrush & \XSolidBrush & \Checkmark   & ~~3(20.90)  & 16(40.09) & 49.43(57)    \\
    \Checkmark   & \XSolidBrush & \Checkmark   & \Checkmark   & \textbf{~~1(39.19)}  & \textbf{~~2(41.77)}  & \textbf{50.57}(59)    \\
    \Checkmark   & \Checkmark   & \Checkmark   & \Checkmark   & ~~1(37.84)  & ~~3(40.00) & 49.16(46) \\
    \bottomrule
    \end{tabular}}
    \end{threeparttable} 
\end{table}

For further validation, we also conducted experiments on a real dynamic video dataset, called UCF101\cite{soomro2012ucf101}. There are $13,320$ video clips in UCF101, the train-test split is: $9,990-3,330$. There is an attempt to convert the RGB-stream dataset to a DVS dataset named UCF101-DVS \cite{bi2020graph} using DAVIS-camera, and their corresponding method RG-CNN achieved $63.2\%$ validation accuracy. For simplicity and excellent compatibility with SNNs, we used the software simulation method in ES-ImageNet\cite{lin2021esimagenet} and generated a new lightweight ES dataset, named ES-UCF101. Compared with the large volume of UCF101-DVS ($28.4GB$ in ZIP format), the volume of our ES-dataset is relatively small, with only $1.66GB$ in ZIP format, which is especially suitable for sparse event recognition algorithm. We set the event ratio close to 0.15 by dynamically adjusting the contrast threshold. Here the $event-ratio = \frac{\#e_{pos}+\#e_{neg}}{H\times W}$, where $\#e_{pos}$ and $\#e_{neg}$ denotes the number of positive and negative events in each event frame, respectively. We also use {\bf pipe-D} here, and $T$ is selected as 15. More details can be found in supplementary materials.

\begin{figure}[!t]
    \centering
    \includegraphics[width=1 \linewidth]{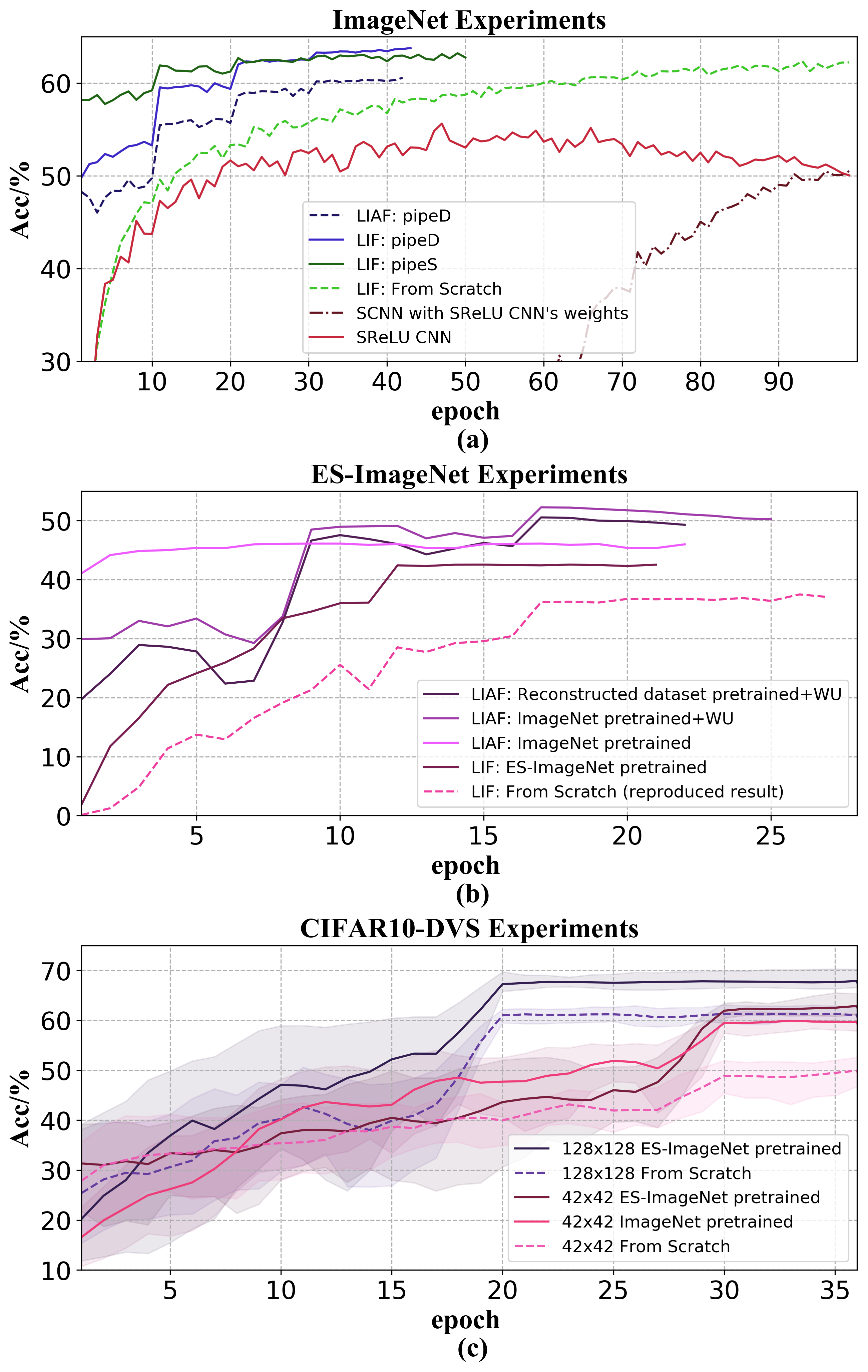}
    \caption{{\bf Test accuracy curves for all experiments.} The baseline of the direct training results is drawn with dashed lines. The shadow in (c) is the fluctuation range of validation accuracy due to different random seeds, and the moving averages of multiple trials (the solid lines) are used to represent the trend.}
    \label{fig:Pipes}
\end{figure}

{\bf Training Result.} We implement a video classification algorithm on UCF101 for the ANN case. One CNN encoder (CNN-Enc) for feature extraction and one RNN (LSTM) decoder (RNN-Dec) for integrating temporal information is adopted, which is similar to LRCN\cite{donahue2015long}. In order to alleviate the disturbance of other factors, only RGB images or event frames are fed into the networks, rather than integrating optical flow, which would degrade our overall experimental results. In this experiment, LIF-SNN and LIAF-SNN act as the spatial-temporal feature encoders only. We firstly pretrain the ANN encoder and decoder on UCF101. And then LIAF-CNN encoders and LSTM decoders are finetuned on ES-UCF101 from pretrained models or scratch. Finally we finetune LIF-CNN encoders and LSTM decoders on ES-UCF101 from pretrained LIAF-CNN encoders and LSTM decoders, as well as from scratch for comparison. We have conducted comprehensive comparative experiments. The efficiency and effectiveness of pretraining from ANNs to SNNs on such a real video dataset could be illustrated by the aforementioned {\bf Table~\ref{table:video-LIAF-results}} and {\bf Table~\ref{table:video-LIF-results}}.

{\bf Analysis.} Results in {\bf Table~\ref{table:video-LIAF-results}} demonstrates that, at the step 1 of {pipe-D}, pretraining CNN-Enc while training RNN-Dec from scratch could accelerate the training and boost the performance of LIAF-Enc on ES-UCF101 by a large margin. Nevertheless, pretraining RNN-Dec, unfortunately, may degrade the results. A possible explanation is that ANNs and SNNs share similar visual feature extraction patterns spatially, but they have different temporal modeling features. More specifically, LIAF and LIF are capable of addressing spatiotemporal features inherently in a collaborative manner, while 2D-CNN can only process visual inputs individually. As {\bf Table~\ref{table:video-LIF-results}} shows, when transferred from LIAF to LIF, pretraining both LIAF-Enc and RNN-Dec performs better, for LIAF and LIF share the similar spatiotemporal mechanism. However, using the RNN-Dec pretrained with the CNN-Enc at step 1 may degrade the final performance, which is consistent with the former results.

\section{Discussion}

{\bf Feature extraction \& inference performance.} The experiments analysis in {Exp. 1} and {Exp. 2} provides some evidence to support the {\bf Hypothesis \ref{Hypothesis:1}}. Similar weights patterns are found in convolution kernels in SNNs and ANNs, which means that although the mechanisms of neurons are different, the preference for spatial features is similar (using BP-based training method), and the features extracted by those convolutional kernels would be similar (some evidence can be found in  {\bf supplementary materials Fig. S3}). The results in {Exp. 3} show that the ability of ANNs and SNNs to represent high-level features may not be similar when dealing with spatial-temporal information streams. This phenomenon emphasizes the importance of intermediate models to smooth the training pipeline, especially when SNNs are trained on dynamic datasets.


{\bf Performance Improvement.} As shown in {\bf Table \ref{table:result-ImageNet} - \ref{table:video-LIF-results}} and the training curves drawn in {\bf Fig. \ref{fig:Pipes}}, the pretrained SNNs can achieve high validation accuracy after only a few training epochs. In Spiking-ResNet18's experiments on ES-ImageNet, after 16 epochs of retraining, the network can obtain 52.25\% validation accuracy, which exceeds all previous experimental records. The LIF-SNN result of 43.74\% is also the highest record for LIF-SNNs. Those results solidly support the {\bf Hypothesis \ref{Hypothesis:2}}. Dozens of repeated experiments are conducted to verify that transfer learning is applicable to LIF-SNN {\bf Fig. \ref{fig:Pipes}} with larger changes in modality and distribution of data than that in the former experiments. As a result, The ANN training speed and the ability of feature encoding are adequately utilized in the pipelines.

In {Exp. 1 and  Exp. 2}, the training of SNN is still improved with the absence of external data. We speculate that previous training methods and structures are not so perfect for giving full play to the capability of SNNs. The inaccurate gradient-estimation caused by the limited expression ability of SNN affects the training progress and restricts the parameter space. However, our pipeline takes advantage of the superior parameter space exploration ability of ANNs. In  Exp. 2, we also introduced external information, which further improves the performance. Although ES-ImageNet is converted from ImageNet, it loses a large amount of color and illumination information. The pretrained models indeed provide some prior knowledge from that information.

{\bf Training Acceleration.} More importantly, the convergence speed of this training pipeline is often several times to tens of times faster than that of the direct training process. In {Exp. 1}, we obtain a LIF-ResNet18 in 30 epochs, where the former record for LIF-ResNet18 needs 300+ epochs to reach that accuracy\cite{fang2021deep}. In experiments on ES-ImageNet, less than 20 epochs are used to achieve the best result, while the compared work uses 50 epochs for training. As for {Exp. 3}, we can even save 45 (2 vs. 47) epochs to get the same performance, which is $>20\times$ faster than the direct training process. Furthermore, the smoothness of training may affect the final results. The introduction of the WU process improves the effect of the pretrain models, which is apparent in {Exp. 2}.

{\bf Limitations.} Limited by the computation resources, we are not able to test the super large-scale pretrained model using this pipeline. Another problem is that the tasks are all CV classification tasks, focusing on feature extraction and high-level features understanding. We hope this technique can be extended to more valuable tasks. In addition, the adjustment of the length of time dimension will also affect the performance, so the pipeline of training is still divided into two steps (for both pipe-S and pipe-D). We hope we can build a complete SNN-ANN transfer learning framework in the future, providing more rigorous theoretical proof to our hypothesis.

\section{Conclusion}

This work combines the SNNs training algorithm and the pretrain technique, which is, to the best of our knowledge, the first attempt to propose a complete pretrain-finetune framework for training LIF-SNNs. The novel training pipeline can significantly accelerate the convergence speed of SNNs' training process and may further improve the validation accuracy with extra data provided, especially when training large-scale SNNs on complicated tasks. 

The pipelines are tested on three different dataset types, including RGB image datasets, converted DVS-datasets, and directly recorded DVS-dataset. On all the above-mentioned datasets, the algorithm achieves higher convergence speeds than direct training. The pipeline provides up to $2.5\times$ and $10\times$ training acceleration on ES-ImageNet and ImageNet-1K with the same or better accuracy. When using external data for pretraining, we achieve SOTA results (52.25\%) on ES-ImageNet. We also provide a time-accuracy benchmark for our newly created ES-UCF101 dataset.

We plan to apply this hybrid-technique training pipeline to other tasks. This will bring a new prospect for SNNs' large-scale and practical applications. The code we used is available in supplementary materials.


\end{document}